\documentclass{article} 
\usepackage{iclr2027_conference,times}

\usepackage{amsmath,amsfonts,bm}

\def\eqref#1{equation~\ref{#1}}

\def\1{\bm{1}}

\DeclareMathAlphabet{\mathsfit}{\encodingdefault}{\sfdefault}{m}{sl}
\SetMathAlphabet{\mathsfit}{bold}{\encodingdefault}{\sfdefault}{bx}{n}

\usepackage{hyperref}
\usepackage{url}
\usepackage{booktabs}   
\usepackage{multirow}   
\usepackage{amsmath}    
\usepackage{amssymb}    
\usepackage{graphicx}   
\usepackage{xcolor}     
\usepackage{wrapfig}
\newcommand{\betterparagraph}[1]{\noindent \textbf{#1. }}

\usepackage{xspace}
\newcommand{\ourmethod}{RaReCache\xspace}

\title{RaReCache: Bridging the Gap in Cross-Model KV Cache Reuse via \underline{Ra}nk disagreement-based Selective \underline{Re}computation}

\author{
    Sreetama Sarkar$^\bigstar$, Saptarshi Mitra$^\blacklozenge$, Sitao Huang$^\blacklozenge$,
    Souvik Kundu$^{\blacktriangle\dagger}$, Peter A. Beerel$^{\bigstar\dagger}$\\
    $^\bigstar$University of Southern California \ \ \ $^\blacklozenge$University of California, Irvine \ \ \ $^\blacktriangle$Intel Labs\\
    \texttt{\{sreetama,pabeerel\}@usc.edu \ \
    \{saptarm,sitaoh\}@uci.edu}\\
    \texttt{\{souvikk.kundu\}@intel.com}
}

\iclrfinalcopy 
\begin{document}

\maketitle
\lhead{Preprint} 

\begingroup
\renewcommand\thefootnote{\fnsymbol{footnote}}
\footnotetext[2]{Equally contributing authors}   
\endgroup

\begin{abstract}
Cross-model KV-cache reuse remains a key challenge in modern LLM serving. Coding agents and multi-model systems increasingly route a shared context across models: a user may switch models mid-session, or a cascade may escalate a difficult query. Because KV caches contain model-specific representations, each switch typically forces the receiving model to prefill the entire context from scratch. Recent work shows that closed-form linear maps can translate KV caches between models in the same family, but transfer accuracy degrades as the model-size gap widens. In this paper, we establish that these transfer failures are concentrated in a small subset of information-dense tokens. To bridge this gap, we introduce \textbf{RaReCache}, a framework that enables a large target model to decode accurately from a cache prefilled by a much smaller source via selective recomputation. RaReCache identifies these critical positions using a novel \emph{rank disagreement} metric, scoring each token by the energy of its mapped KV in output directions weakly supported by the calibration data. 
Across two model families and five benchmarks, on a \(23\times\) parameter gap (Qwen3-0.6B to 14B) recomputing just 30\% of positions retains 95-99\% of the target accuracy, whereas on a \(8.8\times\) gap (Llama3-8B to 70B), recomputing 40\% retains 96.5\% of the target accuracy.
RaReCache largely removes sensitivity to source-model size, and achieves up to a $\mathbf{3.04\times}$ prefill speedup.
For online serving, it handles $1.8\times$ the request throughput of target prefill on a single GPU, and at the target's saturation load, reduces median and 99th-percentile time-to-first-token (TTFT) by $\mathbf{5.0\times}$ and $\mathbf{6.4\times}$ respectively, with a 30\% recompute budget.
RaReCache establishes an efficient serving paradigm where small models prefill on behalf of massive targets, enabling large models to recompute only critical tokens, drastically reducing prefill latency.

\end{abstract}

\section{Introduction}
\label{sec:intro}


Modern large language model (LLM) serving systems increasingly hand off ongoing contexts across models of different capacities. This paradigm underpins cost-quality cascades \citep{chen2023frugalgpt}, dynamic request routing \citep{ong2024routellm}, and coding agents that let users upgrade to stronger models midway through a session. However, each context switch carries a hidden cost. A model's key-value (KV) cache is intrinsically tied to its own architecture and cannot be read by the receiving model, forcing the target model to prefill the entire context from scratch. 

Reusing KV caches is a promising way to avoiding repeated prefill, but existing reuse techniques like prefix caching \citep{kwon2023vllm,zheng2024sglang} or DroidSpeak \citep{liu2026droidspeak} are largely confined to identical architectures. 
Other cross-model approaches train neural fusers or latent adapters to boost accuracy rather than save prefill compute \citep{fu2026c2c,dery2026latentalign}. Closest to our work, \citet{heo2026closedform} introduces a gradient-free, closed-form linear map to convert caches between models within a family. However, their transfer accuracy degrades significantly when the capacity gap between models is large.

To close the accuracy gap in KV transfer, recent reuse strategies rely on partial target-side recomputation. CacheBlend \citep{cacheblend} selectively recomputes tokens based on the $\ell_2$ deviation of their reused KV states in the target's prefix layers. However, 
this requires processing every token through these initial layers, and 
raw error magnitude often fails to predict downstream accuracy loss. Alternatively, DroidSpeak \citep{liu2026droidspeak} proposes selectively recomputing full intermediate layers
which fundamentally requires identical model architectures. Consequently, achieving accurate cross-model KV transfer across large capacity gaps requires a new, architecture-agnostic metric for selective recomputation.

\begin{figure}[t]
    \centering
    \vspace{-5mm}
    \includegraphics[width=\textwidth]{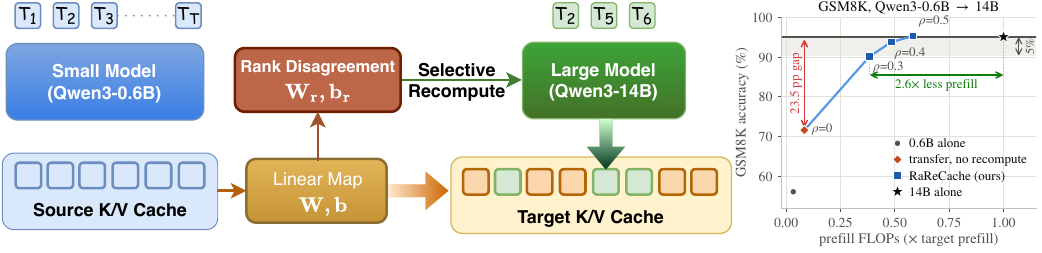}
    \vspace{-8mm}
    \caption{The figure illustrates our \ourmethod pipeline where target cache constitutes mapped cache tokens from source and recomputed tokens in target (left). \ourmethod closes accuracy gap to target while yielding $2.6\times$ lower prefill FLOPs (right).
    }
    \vspace{-5mm}
    \label{fig:intro_figure}
\end{figure}


To address these limitations, we present \textbf{\ourmethod} (\underline{Ra}nk disagreement-based Selective \underline{Re}computation), a framework that bridges the capacity gap in cross-model KV cache transfer. Rather than relying on standard error magnitudes or attention scores, \ourmethod selectively recomputes only the most critical, information-dense tokens in the target model using a novel metric called \emph{rank disagreement}. This approach avoids complete target-side prefilling, enabling highly efficient context hand-offs without sacrificing the reliability of downstream tasks.

\noindent\textbf{Contributions.}
\noindent We introduce \ourmethod, driven by \emph{rank disagreement}, a closed-form metric that identifies which tokens require target-side recomputation without executing any forward passes in the target. By measuring the energy of a token's mapped KV cache outside the well-described calibration subspace, it isolates critical out-of-distribution representations. We demonstrate why common heuristics fail: attention mechanisms select standard formatting tokens the map already reproduces, while raw mapping error conflates harmless variations on template text with crucial novel content. Rank disagreement outperforms both, closing the accuracy gap (\S\ref{sec:exp-selection}).

 We show that a small model's cache, mapped in closed form and selectively repaired using \ourmethod, recovers the accuracy lost by plain transfer even across extreme scale differences. On a \(23\times\) parameter gap (Qwen3-0.6B to 14B), recomputing just 30\% of positions retains 95--99\% of the target's native accuracy on reasoning benchmarks (GSM8K and ARC), largely decoupling transfer quality from the source model's size. These gains generalize seamlessly across architectures: on the Llama 3 family\citep{grattafiori2024llama3}, transferring from the 8B to the 70B model (\(8.8\times\) gap) at a 40\% budget retains 96.5\% of the 70B's native GSM8K performance (\S\ref{sec:exp-recover}).

Because the large target model computes only a fraction of positions selected by rank disagreement, prefill latency during model switch is drastically reduced. We demonstrate that \ourmethod is up to \textbf{3.04}\(\times\) faster than the large model prefilling the full context from scratch (\S\ref{sec:runtime}). 
Under live serving conditions with dynamic request batching, this efficiency reduces head-of-line blocking, lowering median TTFT by $\mathbf{5.0\times}$ and 99th-percentile tail latency (p99) by $\mathbf{6.4\times}$ at a $30\%$ recompute budget.


%

\section{Motivational Case Study}
\label{sec:motivation}
\subsection{The Capacity Bottleneck of Linear Mapping}
\begin{wrapfigure}{r}{0.52\textwidth}
    \vspace{-3mm}
    \centering
\includegraphics[width=0.94\linewidth]{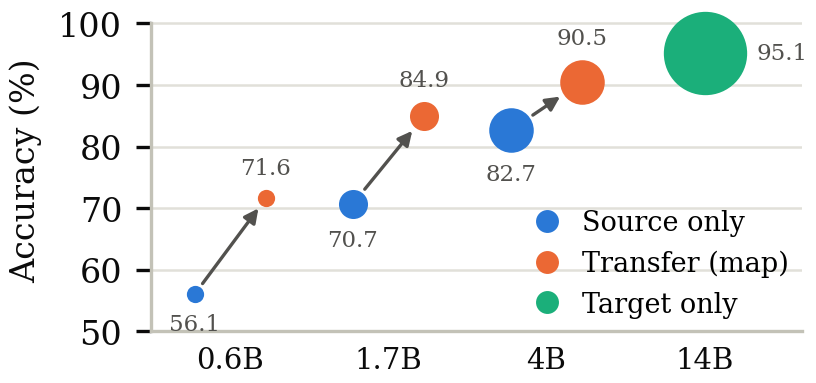}
    \vspace{-5mm}
    \caption{GSM8K acc. for mapping source KV to target on Qwen3. 
    Marker size is proportional to the parameter count of prefill model.
    }
    \label{fig:transfer-pairs}
    \vspace{-5mm}
\end{wrapfigure}

We investigate whether linearly mapping a prefilled KV cache from a small source model, as proposed in \citep{heo2026closedform}, can fully recover a large target model's native accuracy. Figure \ref{fig:transfer-pairs} demostrates mapping Qwen3-0.6B into Qwen3-14B\citep{yang2025qwen3}, a
$23\times$ gap, where marker size is proportional to the parameter count of prefill model. Transfer using linear map yields a GSM8K accuracy of 71.6\% compared to 95.1\% for the 14B model. While mapping lifts the accuracy for every source, the  remaining gap in accuracy between the source and target is still significant, and this gap grows as the source model shrinks. We also observe that the fraction of the target's key variance that the linear map explains barely changes with source size
while the accuracy 
drops a lot.
Mapping more source layers to a single target layer yields negligible gains while increasing the parameter count.

We hypothesize that the mapping extracts what the source representations linearly encode, and thus the shortfall does not necessarily come from a poorly fitted map. The remaining gap is bounded by the small model's representational capacity and must be supplied by the target. We therefore allow the target to \emph{recompute} a fraction of prompt positions while reading the mapped cache from the source model elsewhere.

\subsection{Can Attention identify which tokens to recompute?}
\begin{figure}[t]
    \centering
    \vspace{-5mm}
    \includegraphics[width=0.9\linewidth]{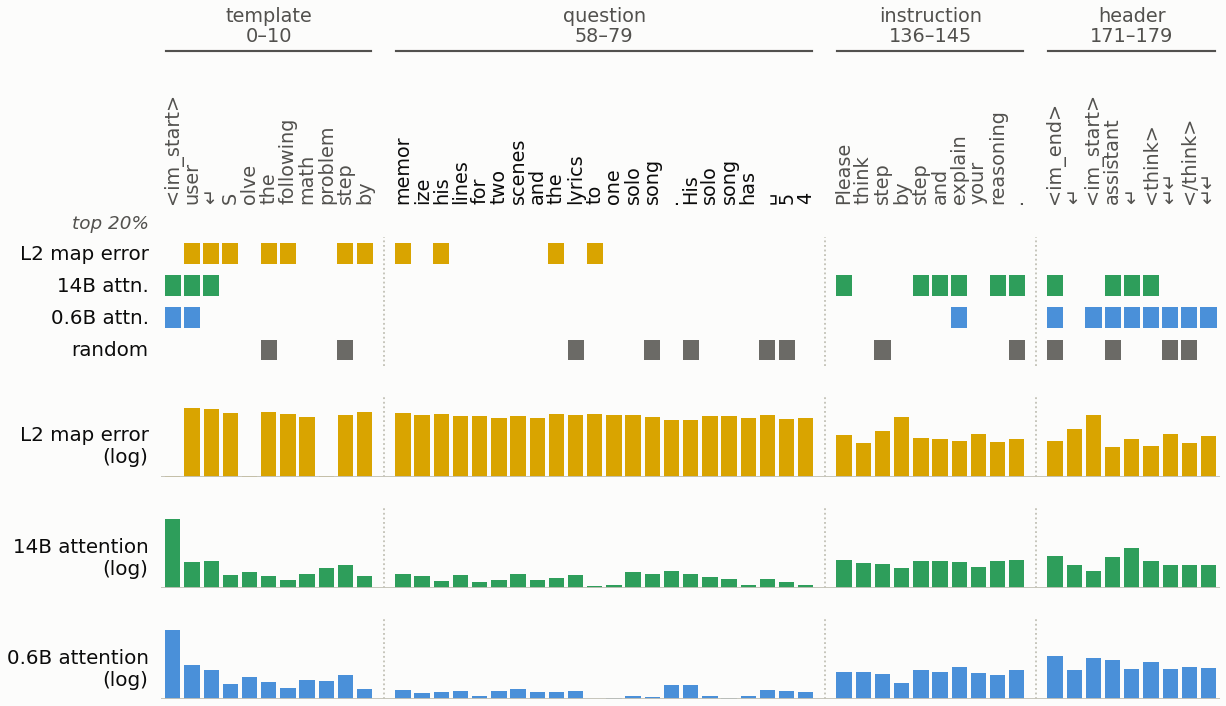}
    \caption{Token-by-token scores for each selector on a single GSM8K prompt. Columns are selected tokens from four sections of the prompt (template, question, instruction, and header). 
    The first four rows marks the tokens in the top 20\% positions determined by four different selection metrics.
    The last three rows are tokenwise scores for three of the four metrics, excluding the random selection metric
    which has no associated score.}
    \vspace{-5mm}
    \label{fig:token-strip}
\end{figure}
\begin{wrapfigure}{r}{0.5\textwidth}
    \vspace{-5mm}
    \centering
    \includegraphics[width=\linewidth]{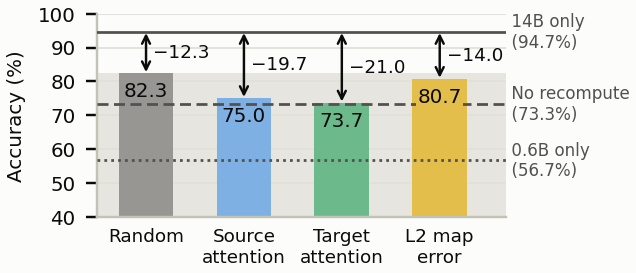}
    \vspace{-8 mm}
    \caption{GSM8K accuracy of
    Qwen3-0.6B$\to$14B when 30\% of prompt positions are recomputed, chosen by four 
    selectors. Arrows give the remaining gap to the 14B alone.}
    \label{fig:selector-gap}
    \vspace{-10pt}
\end{wrapfigure}

We analyze token selection strategies for selective recomputation in the target.
KV-cache eviction methods rank tokens by the  attention they receive, typically computed over a local query window \citep{h2o,snapkv}. We investigate whether attention can be a reliable metric for selective recomputation. Figure~\ref{fig:token-strip} follows one GSM8K prompt, mapped from Qwen3-0.6B into Qwen3-14B, token by token. It shows the map's relative \(\ell_2\) error ($\lVert\mathrm{mapped}-\mathrm{true}\rVert^2/\lVert\mathrm{true}\rVert^2$) calculated for keys and values together, averaged over layers, the source and target model attention each token receives from both models (last 32 query positions for the last quarter of layers), and random selection. In Figure \ref{fig:selector-gap}, we plot the GSM8K accuracy of Qwen3-0.6B$\to$14B on 300 samples when 30\% of prompt positions are recomputed, chosen by each selector.

We observe that attention is overwhelmingly concentrated on the initial sink tokens and prompt-terminating chat templates. The 14B puts 72\% of its attention on the \texttt{<|im\_start|>} sink and 24\% on the rest of the template and instruction, leaving 4\% for the question. The source attention tracks the target model's attention
closely. Both attention selectors score below random and match transfer with no recomputation. Selecting by attention spends its budget on the template, and thus recomputes the tokens the map already reproduces. At $\rho{=}0.2$, attention does not pick a single question token. 
The  \(\ell_2\) mapping-error selector, which reads the 14B's own prefill, also does no better than random, confirming that raw \(\ell_2\) magnitude fails to identify the critical tokens that actually degrade downstream accuracy. This motivates the need for a new metric that focuses on the actual question content and can accurately identify which tokens to recompute.

\section{Methodology}
\label{sec:method}



\begin{figure}[t]
    \centering
    \vspace{-5mm}
    \includegraphics[width=\linewidth]{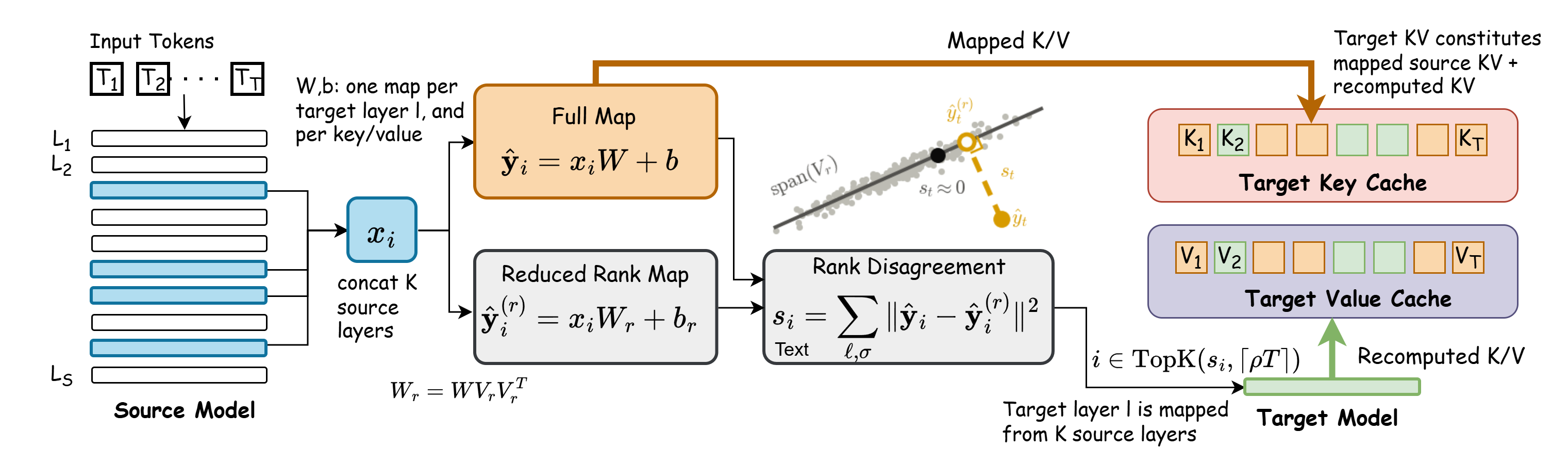}
    \vspace{-8mm}
\caption{The RaReCache Pipeline: The source model prefills the prompt, concatenating $K$ selected layers into input $x_i$. An offline-fitted linear map generates both a full-rank target cache prediction ($\hat{y}_i$) and a reduced-rank projection ($\hat{y}_i^{(r)}$). Rank disagreement ($s_i$) calculates the squared gap between these predictions. As illustrated, tokens whose predictions project outside the familiar calibration subspace $\text{span}(V_r)$ yield high $s_i$ scores. To save compute, the target model only runs a forward pass on the top $\lceil \rho T \rceil$ positions, $\rho$ being the recompute ratio, with the highest $s_i$, merging these recomputed KV tensors with the directly mapped ones to form the final target model cache.}

\vspace{-3mm}
\label{fig:kv_transfer}
\end{figure}

We propose \ourmethod, a framework that bridges cross-model capacity gaps by translating source KV caches and selectively recomputing only the most critical target tokens, as illustrated in Figure~\ref{fig:kv_transfer}. In this section, we formalize the base linear mapping (\S\ref{sec:method-map}), our selective recomputation framework (\S\ref{sec:method-recompute}), and the target-free score used to select those positions (\S\ref{sec:method-rd}).
\subsection{Mapping the cache}
\label{sec:method-map}
Source and target caches differ in depth, width, and number of layers, but within a family they share a tokenizer, so a prompt of $T$ tokens produces $T$ cache entries in both. Following \citet{heo2026closedform}, we fit an independent linear map per target layer $\ell$ and per cache tensor $\sigma \in \{\text{key},\text{value}\}$. For simplicity, we omit $\ell$ and $\sigma$ from the notations below.

\noindent\textbf{Inputs and outputs.}
Let a target model's cache entry be a row vector $\mathbf{y} \in \mathbb{R}^{d_t}$, where $d_t = n^{t}_{\text{kv}} d^{t}_{h}$ concatenates the target's KV heads. To construct the corresponding source input $\mathbf{x} \in \mathbb{R}^{d_s}$, we concatenate the entries from $K$ specific source layers. To determine these $K$ layers, we fit a single-source least-squares regression from each source layer's cache entry to the target layer's, evaluating each by its $R^2$ score, and select the top $K$ source layers with the highest explained variance \citep{heo2026closedform}. This results in an input dimension of $d_s = k\, n^{s}_{\text{kv}} d^{s}_{h}$.
Keys are de-rotated before fitting and re-rotated after mapping, so the map sees content without position and transfers across context lengths.

\noindent\textbf{Fitting.}
Stacking $N$ calibration tokens yields matrices $X \in \mathbb{R}^{N \times d_s}$ and $Y \in \mathbb{R}^{N \times d_t}$. After centering both by subtracting their respective column means $\bar{\mathbf{x}}$ and $\bar{\mathbf{y}}$, we obtain the mapping parameters via ridge regression in closed form:
\begin{equation}
W = (X^{\top}X + \lambda I)^{-1} X^{\top} Y,
\qquad
\mathbf{b} = \bar{\mathbf{y}} - \bar{\mathbf{x}} W,
\qquad
\hat{\mathbf{y}} = \mathbf{x} W + \mathbf{b},
\label{eq:ridge}
\end{equation}
where $W \in \mathbb{R}^{d_s \times d_t}$ is the projection matrix and $\hat{\mathbf{y}}$ is the estimated target cache entry. The bias $\mathbf{b}$ aligns the source and target means. Because the $K$ selected source layers are highly correlated, the $L_2$ penalty $\lambda$ is required to regularize the near-singular $X^{\top}X$. This closed-form fit is highly efficient: a single offline pass over the calibration set accumulates $X^{\top}X$ and $X^{\top}Y$, yielding $W$ via one linear solve without gradient descent.

\subsection{\ourmethod: Selective Recomputation via Rank Disagreement}
\label{sec:method-recompute}
Let $\rho \in [0,1]$ be a recompute budget. We select the top $\lceil \rho T \rceil$ tokens in the prompt to be recomputed by the target model, while reading the linearly mapped cache everywhere else. To maximize efficiency, we use a single shared set of marked positions across all layers (the top 30\% positions at two different layers overlap by 0.87, as shown in the Appendix). 

\noindent\textbf{Rank disagreement.}
\label{sec:method-rd}
Instead of estimating mapping error, we measure how much a token's representation relies on directions the calibration data left undetermined. When the fitted map is applied to the calibration tokens, their predictions $\hat{\mathbf{y}}_1, \dots, \hat{\mathbf{y}}_N$ form a distribution in $\mathbb{R}^{d_t}$. Its covariance $\Sigma_{\hat{y}} = V \Lambda V^{\top}$ yields orthonormal eigenvectors $\mathbf{v}_1, \dots, \mathbf{v}_{d_t}$ and eigenvalues $\lambda_1 \ge \dots \ge \lambda_{d_t}$. The leading directions represent the primary axes of 
variances in the data; along the non-dominant directions, the calibration set barely varies. We call the span of the first $r$ directions the \emph{well-described subspace}, denoted by the columns of $V_r$.

\noindent\textbf{Recomputation score.}
Let $W_r = W V_r V_r^{\top}$ be the map restricted to this subspace, with its bias $\mathbf{b}_r$ refit as in reduced-rank regression \citep{izenman1975}, yielding prediction $\hat{\mathbf{y}}^{(r)}_i$ for token $i$. Rank disagreement is the squared gap between the full-rank and reduced-rank predictions, summed over every mapped layer $\ell$ and tensor type $\sigma$:
\begin{equation}
s_i = \sum_{\ell,\,\sigma}
\bigl\lVert \hat{\mathbf{y}}_i - \hat{\mathbf{y}}^{(r)}_i \bigr\rVert^2,
\qquad
\hat{\mathbf{y}}^{(r)} = \mathbf{x} W_r + \mathbf{b}_r,
\label{eq:score}
\end{equation}
Because the two maps agree entirely within the dominant subspace, their gap captures what the token draws from the remaining, unconstrained directions. 
We recompute the $\lceil \rho T \rceil$ positions with the largest $s_i$.

A token scores near zero when its mapped entry falls safely within the familiar calibration space, regardless of its raw error. Text that recurs in every calibration prompt, such as chat templates or task instructions, defines these leading directions and scores near zero. Conversely, a token scores high when it projects into directions the calibration set rarely explored. Content specific to one item, such as unique entities, quantities, and relations has no counterpart in the calibration data and therefore scores high. Figure \ref{fig:token-strip-with-ours} (Appendix \ref{sec:appendix_results}) demonstrates that this distinction, rather than the magnitude of the error, dictates which recomputations successfully recover downstream accuracy.


\paragraph{Cost.}
The basis $V_r$ is computed once per map from the calibration set, using the source model
and the fitted map but never the target. At inference the map already produces
$\mathbf{z}_t$, and Eq.~\ref{eq:score} can be evaluated as
$\lVert \mathbf{z}_t \rVert^2 - \lVert \mathbf{z}_t V_r \rVert^2$ (derivation in the Appendix \ref{sec:appendix_rd}), one
$d_t \times r$ projection per layer and side. 
To prevent mapping and selection overheads from dominating the runtime, we 
reuse de-rotated source layer blocks across target layers (see \S\ref{sec:phase_opt} for implementation details and hardware profiling).

\section{Experiments}
\label{sec:experiments}
We evaluate \ourmethod across varying capacity gaps, model architectures, and downstream tasks. Specifically, our experiments address three primary research questions:
\begin{enumerate}
    \item[\textbf{Q1}] How much of the target model's native accuracy can RaReCache recover, and how robust is this recovery as the capacity gap widens? (\S\ref{sec:exp-recover})
    \item[\textbf{Q2}] How does rank disagreement compare to standard token selection heuristics, and which tokens actually require target-side recomputation? (\S\ref{sec:exp-selection})
    \item[\textbf{Q3}] Does \ourmethod generalize to the reverse setting (large-to-small transfer), and can it enhance the capabilities of smaller, constrained models? (\S\ref{sec:exp-reverse})
\end{enumerate}

\subsection{Setup}
\label{sec:exp-setup}

\noindent\textbf{Models.}
We evaluate \ourmethod across two model families. In Qwen3, we consider Qwen3-14B as the target and Qwen3-0.6B, 1.7B and 4B as source models, which are 23$\times$, 8.2$\times$ and 3.5$\times$ smaller than the target. In Llama3 we test two pairs, Llama-3.2-3B-Instruct $\rightarrow$ Llama-3.1-8B-Instruct and Llama-3.1-8B-Instruct $\rightarrow$ Llama-3.3-70B-Instruct (2.7$\times$ and  8.8$\times$ model size gaps respectively). 

\noindent\textbf{Map and recomputation.}
We use the linear mapping explained in \S\ref{sec:method-map} with $\lambda{=}0.01$. Every target layer is mapped with $K{=}8$ source layers for Qwen3 and Llama 3B$\to$8B, and $K{=}20$ source layers for {Llama 8B$\to$70B}.
We calibrate the linear mapping matrices using 128k-350k tokens from the training subset of each task, ensuring calibration is strictly disjoint from evaluation items (per-task calibration pools are detailed in \ref{sec:map_data} in the Appendix). For selective recomputation, we apply rank disagreement with a fixed rank of $r{=}128$ across all model pairs. We sweep the recompute budget $\rho \in \{0, 0.2, 0.3, 0.4, 0.5\}$, where $\rho{=}0$ represents transfer without any target-side recomputation, which reproduces the setting in \cite{heo2026closedform}.

\noindent\textbf{Benchmarks.}
We evaluate our approach on five benchmarks, each on its full test set: \textbf{GSM8K}\citep{cobbe2021gsm8k} ($n{=}1319$, chain-of-thought, up
to 1024 new tokens), \textbf{MMLU-Redux}\citep{gema2025mmluredux} ($n{=}5700$, 57 subjects, 0-shot, direct answer),
\textbf{ARC-Challenge}\citep{clark2018arc} ($n{=}1172$) and \textbf{ARC-Easy} ($n{=}2376$, 0-shot,
chain-of-thought), and \textbf{LongBench-E QA}\citep{bai2024longbench}, 0-4k bucket ($n{=}467$: HotpotQA, 2WikiMQA, MultiFieldQA-en, Qasper, TriviaQA; mean prompt length 4.3k tokens).
We use greedy decoding, and
model generations are evaluated using exact match, with the exception of LongBench-E, which is evaluated via token F1.


\subsection{Q1: How much of the target model's accuracy can \ourmethod recover?}
\label{sec:exp-recover}

\begin{figure}[t]
\centering
\vspace{-4mm}
\includegraphics[width=\linewidth]{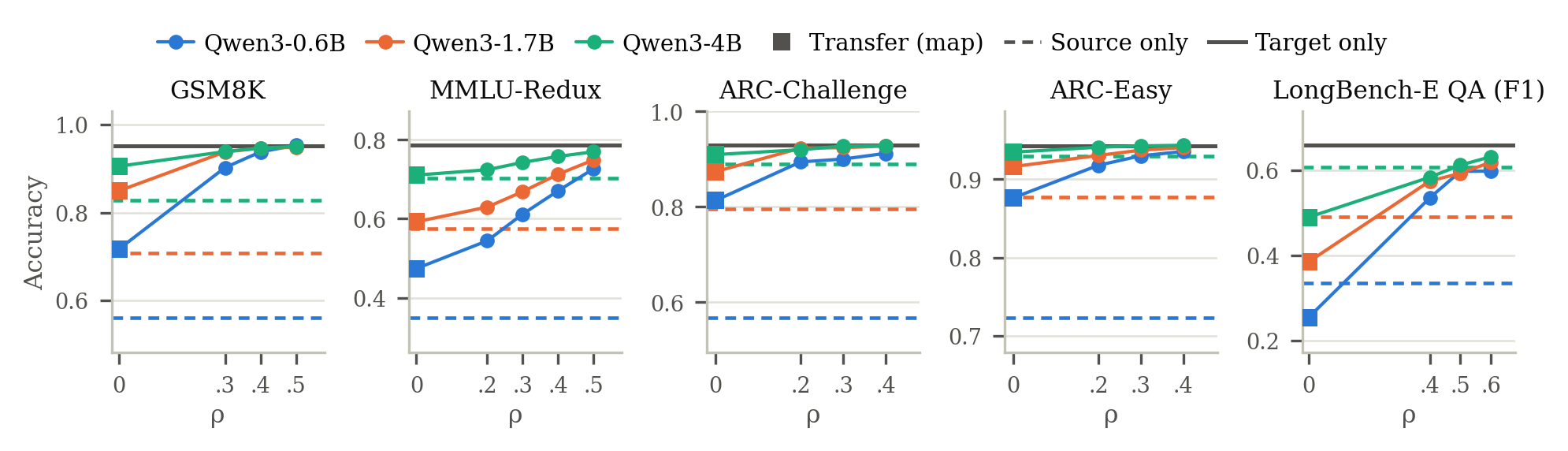}
\vspace{-10 mm}
\caption{
Accuracy or F1 score against the recompute fraction $\rho$ for three Qwen3 sources mapped into Qwen3-14B. Dashed lines are each source answering alone; the solid grey line is the 14B alone. $\rho{=}0$ reproduces results from \cite{heo2026closedform} in our setting.
}
\vspace{-5mm}
\label{fig:by-task}
\end{figure}

We present accuracy against the recompute fraction $\rho$ for three Qwen3 sources mapped into Qwen3-14B in Figure \ref{fig:by-task}. On four of five benchmarks, mapping from a small source into a large target improves accuracy, suggesting that the mapped cache carries information the small model cannot use on its own: plain transfer ($\rho{=}0$) beats the 0.6B alone by 12.4-24.6\,pp (Figure~\ref{fig:by-task}). However, the accuracies as $\rho=0$ are limited by the source model size, with 0.6B model losing 6.7--31.2\,pp to the 14B, retaining 60.3\% on MMLU-Redux and 75.3\% on GSM8K. For
LongBench-E, plain transfer scores 8.2\,F1 below the 0.6B answering alone.

\noindent\textbf{Recomputing a minority of positions recovers most of the gap.}
Recomputing 30\% of positions recovers 94.8\% of the 14B's accuracy on GSM8K (71.6\% to 90.2\%),
96.9\% on ARC-Challenge and 98.6\% on ARC-Easy; on both ARC sets, 20\% already recovers more than 96\% (96.2\% and 97.3\%).
At $\rho{=}0.5$, transfer matches the 14B on GSM8K (95.3\% vs.95.1\%), and reaches 92.4\% on
MMLU-Redux and 90.7\% on LongBench-E. At these points transfer beats the 0.6B alone by
21-39 pp on every benchmark. With a $23\times$ smaller source, recomputing 30\% of
positions recovers more than 94\% of target accuracy on three of five benchmarks, and
recomputing at most 50\% recovers more than 90\% on all five (detailed
numbers are in Table \ref{tab:main} in the Appendix~\ref{sec:appendix_results}).


\noindent\textbf{Recomputation substitutes for source capacity.}
Without recomputation, source size dominates: plain transfer is ordered by source size on
every task (Figure~\ref{fig:by-task}), with a spread across the three sources of 18.9\,pp on GSM8K, 23.6 on MMLU-Redux, 9.7 on ARC-Challenge, 5.9 on ARC-Easy and 23.6\,F1 on LongBench-E.
Recomputation collapses the ordering. On GSM8K the three sources lie within 0.9\,pp at
$\rho{=}0.4$ and 0.5\,pp at $\rho{=}0.5$, on ARC-Challenge within 1.6\,pp and on ARC-Easy
within 0.8\,pp at $\rho{=}0.4$. On LongBench-E the spread falls from 23.6 to 1.9\,F1 at
$\rho{=}0.5$, and  
on
MMLU-Redux, the spread narrows \textbf{more than $\mathbf{5}\times$}, from 23.6 to 4.3\,pp. 

Paired comparisons
show the substitution directly. On GSM8K the 0.6B at $\rho{=}0.3$ matches the 4B with no
recomputation, as it does on ARC-Easy and on ARC-Challenge at $\rho{=}0.4$. On LongBench-E the 0.6B at $\rho{=}0.4$ beats the 4B with no recomputation by 4.6\,F1.
On MMLU-Redux the 0.6B at $\rho{=}0.5$ matches the 4B at $\rho{=}0.2$ while beating the 4B answering alone by 2.5\,pp.


\begin{table}[t]
\centering
\vspace{-3mm}
\caption{\textbf{GSM8K Evaluation on Llama 3:} Accuracy (\%) against the recompute
fraction $\rho$, retention relative to the target is in parentheses. 
The 8B$\to$70B accuracies are reported on the first 300 test items.
}
\label{tab:llama}
\small
\setlength{\tabcolsep}{4pt}
\begin{tabular}{@{}lccccccc@{}}
\toprule
& & & \multicolumn{4}{c}{Transfer, recompute fraction $\rho$} & \\
\cmidrule(lr){4-7}
Pair ($n$) & Gap & Source alone & $0$ & $0.2$ & $0.3$ & $0.4$ & Target alone \\
\midrule
3.2-3B $\to$ 3.1-8B (1319) & $2.5\times$ & 82.3 & 81.7 {\scriptsize(95.2)} & 85.7 {\scriptsize(99.8)} & 84.2 {\scriptsize(98.1)} & \textbf{86.4} {\scriptsize(100.6)} & 85.8 \\
3.1-8B $\to$ 3.3-70B (300) & $8.8\times$ & 82.7 & 81.0 {\scriptsize(85.0)} & 85.3 {\scriptsize(89.5)} & 88.3 {\scriptsize(92.7)} & \textbf{92.0} {\scriptsize(96.5)} & 95.3 \\
\bottomrule
\vspace{-6mm}
\end{tabular}
\end{table}

\noindent\textbf{Evaluation on Llama3.}
We demonstrate results with Llama 3 in Table~\ref{tab:llama} and observe that LLama shows similar behavior as Qwen3. Mapping the 8B into the 70B, an
$8.8\times$ gap, plain transfer scores 81\%, no better than the 8B answering alone, so the mapped cache on its own gives no reason to invoke the 70B. Recomputing
40\% of positions raises it to 92\%, 96.5\% of the 70B, and $\sim$10\,pp above the 8B alone. The 3B$\to$8B pair are close in model size gap, with the 8B only 3.6
points above the 3B. Mapping again fails to beat the source, whereas recomputation matches the 8B (85.7\% vs. 85.8\%) at $\rho{=}0.2$ only.


\subsection{Q2: How does rank disagreement compare to standard token selection heuristics?}
\label{sec:exp-selection}
\begin{wrapfigure}{r}{0.57\textwidth}
    \centering
    \vspace{-1em}
    \includegraphics[width=\linewidth]{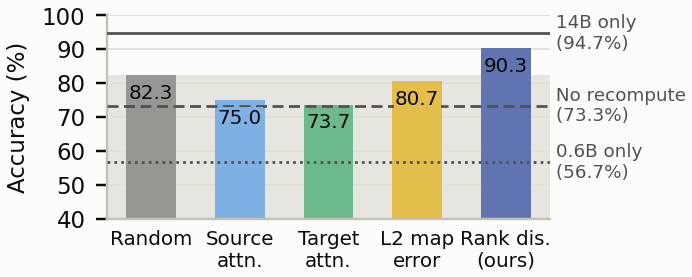}
    \vspace{-8mm}
    \caption{\textbf{Selectors at a matched budget} ($\rho{=}0.3$), for 300 samples on GSM8K, transfer mapping 
    Qwen3-0.6B$\to$14B. The shaded band marks random selection. 
    }
    \label{fig:selector-bars}
    \vspace{-4mm}
\end{wrapfigure}

We compare selectors at a fixed 30\% budget on 300 GSM8K items, holding the map and the
mapped cache fixed so that only the choice of positions changes (Figure~\ref{fig:selector-bars}).
Two of the selectors, the 14B's attention, and the positions with the largest mapping error, are oracles that read the 14B's own prefill and therefore cannot be deployed, but are used for the purpose of our analysis.


\noindent\textbf{Which positions matters more than how many.}
Recomputing 30\% of positions at random closes 42\% of the gap to the 14B (73.3 to 82.3).
Rank disagreement, at the same budget, closes 79\% of it (90.3), and beats every other
selector, including both oracles. We illustrate the selected tokens using rank disagreement in Figure~\ref{fig:token-strip-with-ours} (Appendix \ref{sec:appendix_results}).
Rank disagreement separates the two kinds of error that the $\ell_2$ error magnitude conflates: interpolation error on tokens the calibration set covered, which can be large and costs nothing, and out-of-distribution token errors. Template tokens lie inside the well-described subspace whether or not their position was sampled, so they score near zero even where their $\ell_2$ error is large (Figure~\ref{fig:token-strip-with-ours}). Across the full GSM8K test set, 0.23\% of the positions it recomputes fall in the shared template prefix. The question's content words score high, and they receive under 2\% of the 14B's attention, which is why attention-based selectors are ineffective. 

\subsection{Does \ourmethod generalize to the large-to-small transfer?}
\label{sec:exp-reverse}
A cascade may also hand a large model's context down to a small one. We map the 14B's cache into each of the three smaller models on MMLU-Redux ($n{=}5700$, same calibration recipe as the forward direction), and compare against each small model prefilling for itself (Table~\ref{tab:reverse}).

\begin{table}
\centering
\caption{\textbf{Large-to-small transfer}, Qwen3-14B $\to$ each smaller model on MMLU-Redux. \emph{Alone} is the small model prefilling its own cache. Lift is transfer minus alone.}
\label{tab:reverse}
\small
\begin{tabular}{@{}lccccc@{}}
\toprule
Target & Ratio & Alone & Transfer ($\rho{=}0$) & Lift & Transfer ($\rho{=}0.2$) \\
\midrule
Qwen3-0.6B & $23\times$  & 35.0 & 27.2 & $-7.8$ & 27.9 \\
Qwen3-1.7B & $8.2\times$ & 57.5 & \textbf{71.9} & $\mathbf{+14.4}$ & 71.4 \\
Qwen3-4B   & $3.5\times$ & 70.2 & \textbf{73.3} & $+3.1$ & 73.2 \\
\bottomrule
\end{tabular}
\end{table}

\noindent\textbf{The small model retains its own accuracy, and gains where it has the capacity to
use a richer cache.}
At $8.2\times$ and $3.5\times$, decoding from the 14B's mapped cache beats the small model
decoding from its own prefill, by 14.4\,pp for the 1.7B and 3.1\,pp for the 4B. The 1.7B
result is the largest lift over standalone accuracy in either
direction: the mapped cache carries more than the 1.7B could have encoded itself, and the
1.7B's decoder can exploit it. At $23\times$ the direction fails, losing 7.8\,pp, mirroring
the forward-direction cliff from the other side.

\noindent\textbf{Recomputation does not help in reverse.}
At $\rho{=}0.2$, accuracy moves by at most 0.7\,pp at any ratio, inside noise at this sample
size, where the same budget buys 6--9\,pp forward. Rank disagreement finds positions where
the \emph{source's} cache falls outside the calibration fit; in reverse the source is the
14B, whose cache the map reconstructs well throughout ($R^2_K$ 0.90--0.91 for all three
targets), so there is little of that instability left to find. Whatever bounds the 0.6B's
reverse accuracy is not visible to this score.

\subsection{Comparison with existing reuse strategies}
\label{sec:exp-baselines}
    

\begin{wrapfigure}{r}{0.4\textwidth}
  \vspace{-12pt}
  \centering
  \includegraphics[width=\linewidth]{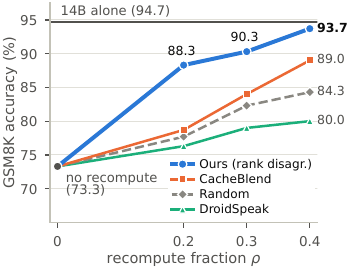}
  \vspace{-16pt}
  \caption{Comparison against existing selective recomputation strategies for Qwen3-0.6B$\to$Qwen3-14B for different recompute ratios for 300 GSM8K samples.}
  \label{fig:selective_recompute}
  \vspace{-6mm}
\end{wrapfigure}

Because existing reuse systems are not designed for cross-architecture size gaps, we adapt their core selection rules for our pipeline and compare them at a matched compute budget \(\rho\) (the fraction of a full target prefill). DroidSpeak \citep{liu2026droidspeak} selectively recomputes layers; since hidden states are incompatible across different model sizes, we adapt this as a layer-prefix recomputation (evaluating the first \(L\) layers). CacheBlend \citep{cacheblend} selects tokens based on \(\ell_2\) KV deviation; we evaluate this in its strongest possible form as an oracle that ranks tokens by their true \(\ell_2\) mapping error against the target's actual cache. 

As shown in Figure~\ref{fig:selective_recompute}, rank disagreement significantly outperforms both adapted baselines. At a 30\% budget (\(\rho=0.3\)) on GSM8K, our method achieves 90.3\% accuracy, outperforming the \(\ell_2\) error oracle (80.7\%) and layer-wise recomputation (79.0\%) by roughly 10 and 11 points respectively. Layer prefixing scales poorly because mapping errors are concentrated in a minority of specific tokens across all depths, rather than isolated to early layers. Meanwhile, the true mapping error oracle performs no better than random selection (82.3\%), confirming that raw \(\ell_2\) magnitude fails to identify the critical tokens that actually degrade downstream reasoning.


\subsection{System Analysis: End-to-End Inference Performance}
\label{sec:runtime}


While the preceding sections establish the recomputation budget $\rho$ from an accuracy perspective, we now evaluate the resulting inference-time speedups and runtime overheads.
%
We benchmark the \ourmethod pipeline for Qwen3-0.6B$\to$14B on a NVIDIA A100 GPU (80GB) with PyTorch 2.6 eager mode on GSM8K (mean prompt length $T{=}149$).
We measure the latency across three sequential stages: projecting the source KV cache via the linear map (along with tensor preprocessing and feature gathering), scoring positions via rank disagreement, and executing the target model's forward pass over the selected $\lceil \rho T \rceil$ positions.
%
%
We compare against the target model prefilling the entire prompt in all experiments as a baseline.
%
Our latency evaluation assumes the source model has already prefilled the prompt, reflecting common deployment scenarios such as model cascades or multi-turn agents where the smaller model runs first (same setup as \cite{heo2026closedform}) and its KV cache is already available.
%
%
Unless otherwise stated, all reported latencies reflect GPU-busy time: the sum of individual GPU kernel durations.
%


\begin{figure}[t]
\centering
\includegraphics[width=\textwidth]{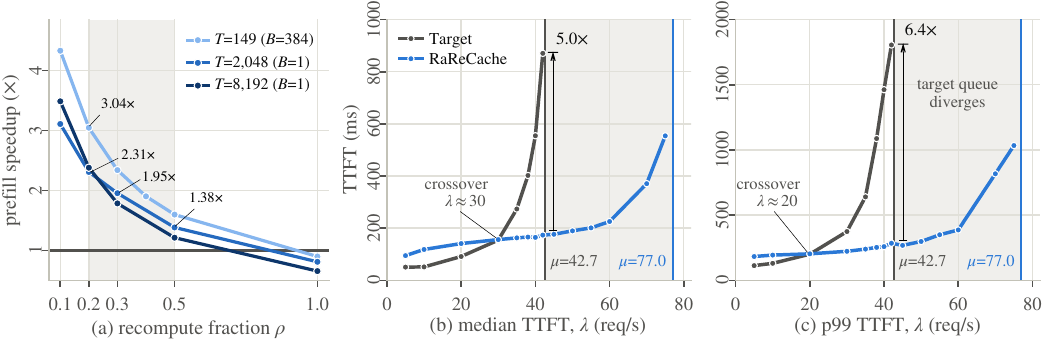}
\vspace{-6mm}
\caption{
(a)~Prefill speedup across recompute fractions $\rho$ (from \S\ref{sec:exp-recover} shaded); 
(b, c)~Median and p99 TTFT vs request rate under Poisson arrivals ($T{=}149$, $\rho{=}0.3$). 
The shaded region shows the divergence of the target prefill queue while \ourmethod operates  at lower device utilization.
}
\vspace{-5mm}
\label{fig:single_batch}
\end{figure}

\betterparagraph{Impact of Recomputation Budget on Runtime}
In Figure ~\ref{fig:single_batch}a, we compare prefill speedup against recomputation fraction $\rho$. Overall latency gain is determined by $\rho$, meaning lower recomputation results in greater speedup.
But this latency savings do not scale linearly with $\rho$.
For instance, recomputing at $\rho{=}0.3$ requires $38\%-63\%$ of the full prefill latency, not $30\%$. 
Two architectural bottlenecks not captured by theoretical FLOP counts explain this gap:
\textit{(a)} Loading the target model’s 27.5GB of weights from high-bandwidth memory is required, whether recomputing a single position or the entire sequence, which dominates execution time at shorter prompt lengths, and
\textit{(b)} Every recomputed query token must still attend all $T$ prompt keys, not just the recomputed subset.
%
%
%
%
%
For a single request of $T{=}2048$, we achieve speedups of $\mathbf{1.38\times}$, $\mathbf{1.95\times}$ and $\mathbf{2.31\times}$  at $\rho=0.5, 0.3, 0.2$ respectively.
$\rho \in [0.2, 0.3]$ is optimal, capturing most prefill latency savings while maintaining near-native target accuracy ($\rho=0.3$ recovers 95-99\% of target model accuracy, as observed in Figure \ref{fig:by-task}).
These latency savings widen under request batching, because native target prefill saturates compute at a batch of 4 ($6,256$ tokens/s) while our smaller compute kernels achieve higher hardware occupancy, throughput nearly doubles to $12,378$ tokens/s, pushing speedups up to $\mathbf{3.04\times}$ and $\mathbf{2.36\times}$ at device saturation for $\rho{=}0.2$ and $0.3$ respectively with T=149 (see \S\ref{sec:runtime-throughput} for complete batch sweeps).

\betterparagraph{Queuing Dynamics and TTFT Under Load}
Figure~\ref{fig:single_batch}(b,c) demonstrates an interactive serving scenario, where user-perceived TTFT is the sum of queuing delay and prefill execution latency.
Accelerating prefill reduces both terms simultaneously: requests complete faster, maintaining lower compute utilization and preventing queue accumulation. 
Evaluating both systems on a live server (\S\ref{sec:runtime-queueing} details benchmarking setup and scheduler configuration) under identical Poisson arrival traces shows that \ourmethod significantly flattens the TTFT curve under load.
In this specific experiment,  we measure end-to-end wall-clock time to account for live queuing delays and framework dispatch overhead.
The requst rate ($\lambda$ req$/$s) increased eightfold, causing the median TTFT of \ourmethod to rise from $95$ ms to $173$ ms. Whereas, the baseline target prefill surged from $51$ ms to $870$ ms.
At $\lambda{}=42$ req$/$s, the saturation threshold ($\mu$) of the target, \ourmethod reduces median TTFT by $\mathbf{5.0\times}$ and p99 latency by $\mathbf{6.4\times}$.

Tail latency benefits appear earlier, crossing the baseline happens at $\lambda \approx 20$ requests/s, compared to $\lambda \approx 30$ req/s for the median.
The target forms larger batches under contention,
which inflates head-of-line blocking for subsequent arrivals.
%
%
The baseline queue diverges between $42.7$ and $77.0$ req$/$s, while \ourmethod operates comfortably at 
lower device utilization.
%
To serve this range of request rates, the baseline needs at least two GPUs provisioned, whereas we only need one.
%
Under light traffic ($\lambda < 30$ requests/s), requests process mostly at batch size 1. There the target prefill is faster because \ourmethod incurs eager-mode framework dispatch overhead across its smaller kernels.

\section{Conclusion}
In this work, we introduced \ourmethod, a framework that bridges the severe accuracy degradation typically seen when transferring KV caches across models with massive capacity gaps. By demonstrating that cross-model transfer failures are heavily concentrated in a minority of information-dense tokens, we developed a novel metric based on rank disagreement that isolates out-of-distribution representations without requiring target-side prefill. \ourmethod allows a massive target model to inherit a context prefilled by a much smaller source model, recovering near-native reasoning accuracy by selectively recomputing as little as 30\% of the prompt. Ultimately, this approach decouples transfer quality from source-model capacity and delivers significant prefill speedups, unlocking highly efficient agentic LLM serving paradigms.
%
We currently report efficiency using standard attention kernels and plan to implement custom sparse causal kernels in the future to attain peak theoretical gains.

\subsection*{AI use statement}
In compliance with ICLR guidelines, we disclose the use of generative AI (large language models) across several phases of this work. For tasks requiring disclosure, AI tools were used to interpret measurement logs, propose and critique hypotheses regarding the effects reported in \S\ref{sec:motivation} and \S\ref{sec:experiments}, and design experiments by suggesting high-value configurations for the sweeps reported herein. For tasks where disclosure is recommended, we utilized these tools to assist with literature search, drafting and editing sections of the manuscript, and writing scripts to generate figures. 
Importantly, the core implementation of our method, including the rank-disagreement scoring, and selective recomputation, was written entirely by the authors without AI assistance.
We did not use AI to generate synthetic datasets or develop theoretical models; the derivation in Appendix~A was written and verified by the authors.
We have reviewed all AI-assisted content and take full responsibility for the final text, claims, and artifacts.

\subsection*{Reproducibility statement}

To ensure reproducibility, we have detailed all experimental setups, model configurations, and evaluation metrics in the manuscript. Evaluations use open-weights models from Qwen3 and Llama 3 families and standard, publicly available benchmarks like GSM8K, MMLU-Redux, ARC-Challenge, ARC-Easy, and LongBench-E QA. 
The system-level inference and queuing benchmarks transparently specify the hardware and software environments.
The complete source code, including the linear mapping implementation, rank disagreement scoring, system optimization and evaluation scripts will be made publicly available upon de-anonymization and publication.




\bibliography{iclr2027_conference}
\bibliographystyle{iclr2027_conference}

\appendix
%

\section{Rank disagreement derivation}
\label{sec:appendix_rd}

This appendix derives the score $\lVert \mathbf{z}_t \rVert^2 - \lVert \mathbf{z}_t V_r \rVert^2$ from Eq.~\ref{eq:score} and states the
behaviour of the score at the two extremes of the probe rank. We work with one map, that
is one target layer and one side; the score sums these independently.

\paragraph{Setup.}
Let $W \in \mathbb{R}^{d_s \times d_t}$ and $\mathbf{b}$ be the fitted map of
Eq.~\ref{eq:ridge}, and let $\bar{\mathbf{x}}$ be the calibration mean of the source
features. The predictions on the calibration set have mean
\begin{equation}
\hat{\boldsymbol{\mu}} = \bar{\mathbf{x}} W + \mathbf{b},
\label{eq:app-mu}
\end{equation}
since the map is affine, and covariance
$\Sigma_{\hat{y}} = \frac{1}{N}\sum_n (\hat{\mathbf{y}}_n - \hat{\boldsymbol{\mu}})^{\top}
(\hat{\mathbf{y}}_n - \hat{\boldsymbol{\mu}}) = V \Lambda V^{\top}$, with $V$ orthonormal
and $\Lambda$ diagonal with entries in decreasing order. Let $V_r$ be the first $r$ columns
of $V$ and
\begin{equation}
P_r = V_r V_r^{\top}
\end{equation}
the orthogonal projection onto the well-described subspace. Being an orthogonal projection,
$P_r$ satisfies $P_r^{\top} = P_r$ and $P_r^2 = P_r$, and $I - P_r$ projects onto the
orthogonal complement.

\paragraph{The rank-$r$ map.}
Reduced-rank regression \citep{izenman1975} restricts the map to the leading directions of
its own fitted values, and refits the bias so the restricted map still sends the average
input to the average output:
\begin{equation}
W_r = W P_r,
\qquad
\mathbf{b}_r = \hat{\boldsymbol{\mu}} - \bar{\mathbf{x}} W_r,
\qquad
\hat{\mathbf{y}}^{(r)}_t = \mathbf{x}_t W_r + \mathbf{b}_r .
\label{eq:app-rrr}
\end{equation}
This is not the truncated SVD of $W$. The truncated SVD keeps the directions in which $W$
itself is largest; reduced-rank regression keeps those in which the map's outputs on real
calibration data vary most. The two coincide only when the source features are isotropic.
Since the $k$ concatenated source layers are strongly correlated, they are far apart here,
and truncating $W$ directly would spend rank on directions the data never excites.

\paragraph{Derivation.}
Write $\mathbf{z}_t = (\mathbf{x}_t - \bar{\mathbf{x}})W$ for the centered prediction.
Substituting Eq.~\ref{eq:app-rrr} into the difference of the two predictions and then
Eq.~\ref{eq:app-mu} for $\hat{\boldsymbol{\mu}}$,
\begin{align}
\hat{\mathbf{y}}_t - \hat{\mathbf{y}}^{(r)}_t
&= (\mathbf{x}_t W + \mathbf{b})
 - \bigl(\mathbf{x}_t W P_r + \hat{\boldsymbol{\mu}} - \bar{\mathbf{x}} W P_r\bigr) \\
&= \mathbf{x}_t W + \mathbf{b}
 - \mathbf{x}_t W P_r - \bar{\mathbf{x}} W - \mathbf{b} + \bar{\mathbf{x}} W P_r \\
&= (\mathbf{x}_t - \bar{\mathbf{x}}) W (I - P_r)
 \;=\; \mathbf{z}_t (I - P_r).
\label{eq:app-diff}
\end{align}
The bias terms cancel, which is why the score does not depend on either model's mean cache.
Because $I - P_r$ is an orthogonal projection and the $\mathbf{v}_i$ are orthonormal,
\begin{equation}
\bigl\lVert \mathbf{z}_t (I - P_r) \bigr\rVert^2
= \sum_{i > r} \langle \mathbf{z}_t, \mathbf{v}_i \rangle^2
= \lVert \mathbf{z}_t \rVert^2 - \lVert \mathbf{z}_t V_r \rVert^2 ,
\label{eq:app-pythagoras}
\end{equation}
which is Eq.~\ref{eq:score} together with the form used to compute it. Summing
Eq.~\ref{eq:app-diff} over layers and sides gives the score of Eq.~\ref{eq:score}. The
right-hand form of Eq.~\ref{eq:app-pythagoras} is the cheap one: the map already computes
$\mathbf{z}_t$, so the score costs one $d_t \times r$ projection per layer and side.

\paragraph{Limits in $r$.}
At $r{=}0$, $P_r = 0$ and $s_t = \sum \lVert \mathbf{z}_t \rVert^2$: the score degenerates
to the size of the centered prediction, which measures how unusual a token is without
reference to how well the map handles it. At $r{=}d_t$, $P_r = I$ and $s_t = 0$ for every
token. Between the two, $s_t$ measures only the energy that falls outside the directions
the calibration set covers. 


\section{System Design}

\subsection{Mapping Implementation and Runtime Efficiency}
\label{sec:phase_opt}
\begin{figure}[t]
\centering
\includegraphics[width=\textwidth]{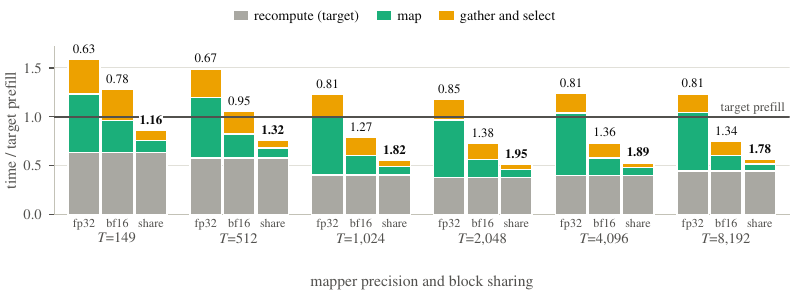}
\caption{\textbf{What \ourmethod costs, relative to the target prefill.}
~Phase-by-phase runtime across implementation variants at increasing context lengths ($T{=}149$  is the mean GSM8K prompt length), with numbers above the bars showing overall speedups. While naive FP32 mapping consumes nearly as much time as a full prefill, BF16 execution combined with source block sharing reduces mapping and selection overhead to a small fraction of total latency.
}
\label{fig:handoff_wide}
\end{figure}

For a Qwen3-0.6B$\to$14B KV transfer experiment on GSM8K with $\rho{=}0.3$, the linear map contains $6.7 \times 10^{8}$ parameters; roughly $5\%$ of the target model's $1.3 \times 10^{10}$ parameters.
This suggests that by arithmetic count alone (excluding partial target prefill), cache translation should cost only a fraction of a full prefill.
But in practice, executing the map in FP32 severely bottlenecks our mechanism because an NVIDIA A100 routes FP32 matmul through vector units at $19.5$ TFLOP/s rather than utilizing its Tensor Cores, which deliver $312$ TFLOP/s in BF16.
Consequently, an unoptimized FP32 map dominates latency, causing cache transfer to run slower than native target prefill ($0.63$-$0.85\times$).
We eliminate this overhead with two systems optimizations that preserve the algorithm:
\begin{itemize}
    \item Executing the map in BF16 to leverage Tensor Cores, raising relative performance to $0.78$-$1.38\times$.
    \item Sharing precomputed source blocks. The reference implementation redundantly reconstructs the same $50$ de-rotated source blocks $640$ times across target layers and KV heads; caching these blocks once per prompt pushes overall speedup to $\mathbf{1.16}$-$\mathbf{1.95\times}$ (Figure~\ref{fig:handoff_wide}, Table~\ref{tab:handoff}).
\end{itemize}

At $T{=}2048$, these optimizations reduce mapping latency from $195$ ms to $27$ ms ($7.2\times$) and token selection from $70$ ms to $17$ ms ($4.2\times$), leaving target-side recomputation as the main source of remaining latency.
%
%
The theoretical speedup ceiling of \ourmethod can be analytically calculated and explained in detail in \S~\ref{sec:ceiling}.

\begin{table}[t]
\caption{\textbf{Prefill speedup over the target baseline, at recomputation budget $\rho{=}0.3$.} Rows are cumulative, each adding an optimization to the row above.
The first three rows are single requests (batch size 1); the last is the
largest batch the device holds, which is $384, 128, 64, 32, 16$ and $8$ requests from left to
right ($\approx$65k tokens in flight). 
}
\label{tab:handoff}
\centering
\small
\begin{tabular}{lcccccc}
\toprule
Prompt length $T$ & 149 & 512 & 1{,}024 & 2{,}048 & 4{,}096 & 8{,}192 \\
\midrule
fp32 map              & 0.63 & 0.67 & 0.81 & 0.85 & 0.81 & 0.81 \\
$+$ bf16 map          & 0.78 & 0.95 & 1.27 & 1.38 & 1.36 & 1.34 \\
$+$ block sharing     & 1.16 & 1.32 & 1.82 & \textbf{1.95} & 1.89 & 1.78 \\
$+$ saturating batch  & \textbf{2.36} & 2.34 & 2.30 & 2.23 & 2.09 & 1.89 \\
\bottomrule
\end{tabular}
\end{table}

\subsection{Batching and Peak Throughput}
\label{sec:runtime-throughput}

\begin{figure}[h]
\centering
\includegraphics[width=\textwidth]{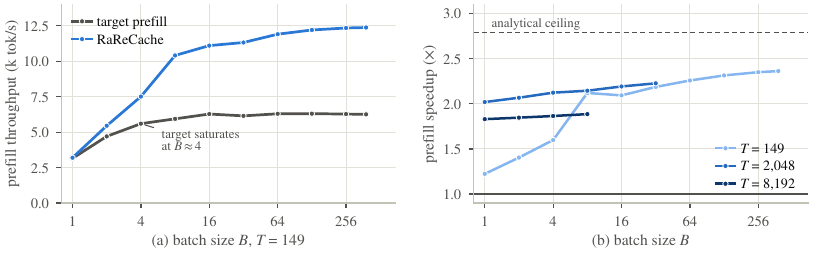}
\caption{
(a) Prefill throughput scaling at $T{=}149$, where target only prefill saturates near batch size $4$ while \ourmethod continues to scale. (b) Prefill speedup as a function of batch size across three prompt lengths; the dashed horizontal line indicates the theoretical speedup ceiling. 
}
\label{fig:batch}
\end{figure}

Figures \ref{fig:batch}a and b illustrate the batch-scaling behavior from Table ~\ref{tab:handoff}.
Native target prefill throughput plateaus beyond an order of magnitude of batch size ($6,256$ tokens/s).
This is because its large GEMMs already saturate GPU compute, leaving no room for amortization.
In contrast, \ourmethod pipeline consists of smaller kernels that underutilize GPU compute at lower batch sizes.
Increasing the batch size improves Streaming Multiprocessor (SM) occupancy, nearly doubling prefill throughput to $12,378$ tokens/s.

\subsection{
Live Serving Evaluation and Queuing Methodology
}
\label{sec:runtime-queueing}


\paragraph{Scheduler.} Both configurations are evaluated using a first-come-first-served (FCFS) dynamic batching scheduler deployed on a live server. Requests arrive via a homogeneous Poisson process at rate $\lambda$. When the GPU becomes idle, the scheduler dispatches up to $32$ queued requests (oldest first) as a single execution batch, leaving any remaining requests queued for the subsequent iteration. Each evaluation sweeps $500$ requests, discarding the initial $50$ to capture steady-state queuing behavior. TTFT is measured in wall-clock time rather than GPU-busy time to capture both queuing delays and framework-level eager dispatch overheads.

\paragraph{Saturation Throughput and Batch Dynamics.} We measure each system's empirical saturation threshold $\mu$ by executing three consecutive maximum-capacity batches ($32$ requests each) back-to-back under a permanently saturated queue:
\begin{equation}
\mu = \frac{3 \times 32}{\text{wall-clock execution time}}.
\end{equation}
This yields saturation throughputs of $\mu_{\text{target}} = 42.7$ and $\mu_{\text{\ourmethod}} = 77.0$\,requests/s ($1.80\times$ capacity advantage, consistent with \S\ref{sec:runtime-throughput}). We report offered load normalized to target capacity, $u_{\text{target}} = \lambda / \mu_{\text{target}}$. Because the baseline queue diverges for $\lambda \geq \mu_{\text{target}}$, the sweep caps at $\lambda = 42$\,requests/s ($u_{\text{target}} = 0.98$).

Under dynamic queuing, the mean batch size $b^*$ satisfies the equilibrium condition $b^* \approx \lambda \cdot S(b^*)$, where $S(b)$ is the batch service latency. Because $S_{\text{target}}(b) \gg S_{\text{\ourmethod}}(b)$, the baseline target experiences severe queue buildup, causing its mean batch size to extend to $24.1$ requests at $\lambda = 42$\,requests/s (approaching the $32$-request cap). In contrast, the shorter service time of \ourmethod clears requests rapidly, keeping its mean batch size at only $5.9$ and drastically preventing head-of-line blocking (Table~\ref{tab:ttft_load}).

\begin{table}[h]
\caption{\textbf{TTFT under Poisson load} ($T{=}149$, $\rho{=}0.3$, max batch size $32$, wall clock). Offered load is normalized to target saturation capacity ($u_{\text{target}} = \lambda / \mu_{\text{target}}$). Speedup ratios represent target $\div$ \ourmethod ($>1\times$ denotes \ourmethod is faster). 
}
\label{tab:ttft_load}
\centering
\small
\setlength{\tabcolsep}{4.5pt}
\begin{tabular}{lcccccc}
\toprule
& & Mean Batch & \multicolumn{2}{c}{p50 TTFT} & \multicolumn{2}{c}{p99 TTFT} \\
\cmidrule(lr){4-5} \cmidrule(lr){6-7}
$\lambda$ (req/s) & $u_{\text{target}}$ & tgt / ours & tgt $\to$ ours (ms) & Ratio & tgt $\to$ ours (ms) & Ratio \\
\midrule
5           & 0.12          & 1.1 / 1.2            & 51 $\to$ 95            & 0.54$\times$          & 113 $\to$ 182           & 0.62$\times$          \\
10          & 0.23          & 1.2 / 1.7            & 51 $\to$ 118           & 0.43$\times$          & 130 $\to$ 193           & 0.67$\times$          \\
20          & 0.47          & 2.1 / 2.8            & 91 $\to$ 141           & 0.65$\times$          & 200 $\to$ 203           & 0.99$\times$          \\
30          & 0.70          & 4.0 / 4.1            & 154 $\to$ 156          & 0.99$\times$          & 374 $\to$ 222           & 1.68$\times$          \\
35          & 0.82          & 7.4 / 4.7            & 274 $\to$ 163          & 1.68$\times$          & 640 $\to$ 239           & 2.67$\times$          \\
40          & 0.94          & 17.0 / 5.5           & 554 $\to$ 165          & 3.37$\times$          & 1462 $\to$ 258          & 5.66$\times$          \\
\textbf{42} & \textbf{0.98} & \textbf{24.1 / 5.9}  & \textbf{870 $\to$ 173} & \textbf{5.02$\times$} & \textbf{1805 $\to$ 284} & \textbf{6.35$\times$} \\
\bottomrule
\end{tabular}
\end{table}

\subsection{Analytical Speedup Ceiling and Attention Scaling}
\label{sec:ceiling}
Once mapper overhead is eliminated, target-side recomputation accounts for $73$--$79\%$ of total \ourmethod latency (\S\ref{sec:phase_opt}).
In our implementation, recomputed query tokens attend over the entire context $T$ via an explicit attention mask, evaluating all $\rho T^2$ query-key pairs rather than only the causally valid subset. Comparing total FLOPs between native target prefill and \ourmethod pipeline yields the speedup ceiling:
\begin{equation}
\frac{\text{Target FLOPs}}{\text{\ourmethod FLOPs}} = \frac{2 P_{\mathcal{T}} T + \frac{1}{2}\alpha T^2}{\left(2P_{\text{map}} + 2P_{\text{probe}} + 2\rho P_{\mathcal{T}}\right) T + \rho\,\alpha T^2}, \qquad \alpha = 4 L_{\mathcal{T}} H_{\mathcal{T}} d_h,
\label{eq:workratio}
\end{equation}
where $P_{\mathcal{T}}$, $P_{\text{map}}$, and $P_{\text{probe}}$ represent the parameter counts of the target model, linear map, and rank-disagreement selector, respectively, and $L_{\mathcal{T}}$, $H_{\mathcal{T}}$, and $d_h$ denote target layer depth, head count, and head dimension.

As context length $T$ increases, quadratic attention FLOPs dominate linear feed-forward terms, causing theoretical speedup to asymptotically approach $1 / (2\rho)$ ($1.67\times$ at $\rho = 0.3$). Consequently, the speedup ceiling declines with prompt length, dropping from $2.74\times$ at $T{=}2{,}048$ to $2.28\times$ at $T{=}32{,}768$.

A dedicated attention kernel that skips causally masked query-key pairs would reduce the recomputation attention cost to $\frac{1}{2}\rho \alpha T^2$, raising the asymptotic ceiling to $1/\rho$ ($3.33\times$ at $\rho = 0.3$). Under this optimized execution, speedups would scale positively with context length (reaching $2.95\times$ at $T{=}32{,}768$), making long-sequence serving the most favorable regime. We report benchmarks using standard, unspecialized attention kernels and leave custom sparse causal implementations to future work.
\section{Experimental Results}
\label{sec:appendix_results}

\subsection{Mapping.}
\label{sec:map_data}
We use the map of \S\ref{sec:method-map} with $\lambda{=}0.01$ throughout, choosing the $K$
source layers for each target layer by $R^2$. Every target layer is mapped for both keys
and values:
\begin{itemize}
    \item \textbf{Qwen3} (all three sources): $k{=}8$, giving $8192 \to 1024$ dimensions per
    token over 40 target layers, for 671M map parameters.
    \item \textbf{Llama 3B$\to$8B}: $k{=}8$, $8192 \to 1024$ over 32 target layers, 537M
    parameters.
    \item \textbf{Llama 8B$\to$70B}: $k{=}20$, $20480 \to 1024$ over 80 target layers,
    3.36B parameters.
\end{itemize}

Calibration data is from the train subset of the same task, formatted similarly as the evaluation prompts, and disjoint from every evaluation item:
\begin{itemize}
    \item GSM8K uses its train split (128k tokens, stride 4; 320k, stride 2, for Llama
    8B$\to$70B).
    \item MMLU-Redux uses MMLU validation (128k, stride 1). MMLU-Redux is a subset of items
    from the MMLU test split, so the two are disjoint.
    \item ARC uses the pooled ARC-Challenge and ARC-Easy train splits (150k, stride 3).
    \item LongBench-E uses a pool drawn from its 4-8k and 8k+ buckets and from LongBench's
    standard split (350k tokens at stride 32). We remove every item that shares a question or a passage of 30 or more words with the 0-4k evaluation bucket.
\end{itemize}
Token positions are subsampled at the stated stride to ensure that the calibration budget covers a diverse range of contexts across the entire corpus. 

\paragraph{\ourmethod Results for Small-to-Large Prefill}
The detailed accuracy and retention numbers for the Figure \ref{fig:by-task} is presented in the Table \ref{tab:main}.

\begin{table}[t]
\centering
\caption{\textbf{Qwen3-0.6B $\to$ Qwen3-14B across five benchmarks.}
Accuracy (\%, F1 for LongBench-E) as a function of the fraction $\rho$ of prompt positions
recomputed in the target. Retention relative to the 14B is in parentheses.
$\rho{=}0$ is plain transfer. $^\dagger$\,Not significantly different from the 14B
($p{=}0.86$). $^\ddagger$\,Below the source model alone. Each recompute
budget improves over $\rho{=}0$. A dash means the arm was not run.}
\label{tab:main}
\small
\setlength{\tabcolsep}{4pt}
\begin{tabular}{@{}lccccccc@{}}
\toprule
& & \multicolumn{5}{c}{Transfer, recompute fraction $\rho$} & \\
\cmidrule(lr){3-7}
Benchmark ($n$) & 0.6B alone & $0$ & $0.2$ & $0.3$ & $0.4$ & $0.5$ & 14B alone \\
\midrule
GSM8K (1319)          & 56.1 & 71.6 {\scriptsize(75.3)} & --- & 90.2 {\scriptsize(94.8)} & 93.8 {\scriptsize(98.6)} & \textbf{95.3}$^\dagger$ {\scriptsize(100.2)} & 95.1 \\
MMLU-Redux (5700)     & 35.0 & 47.4 {\scriptsize(60.3)} & 54.4 {\scriptsize(69.3)} & 61.2 {\scriptsize(77.9)} & 67.1 {\scriptsize(85.4)} & \textbf{72.6} {\scriptsize(92.4)} & 78.6 \\
ARC-Challenge (1172)  & 56.7 & 81.3 {\scriptsize(87.5)} & 89.4 {\scriptsize(96.2)} & 90.0 {\scriptsize(96.9)} & \textbf{91.2} {\scriptsize(98.2)} & --- & 92.9 \\
ARC-Easy (2376)       & 72.3 & 87.6 {\scriptsize(92.9)} & 91.8 {\scriptsize(97.3)} & 93.0 {\scriptsize(98.6)} & \textbf{93.6} {\scriptsize(99.2)} & --- & 94.3 \\
LongBench-E QA (467)  & 33.5 & 25.4$^\ddagger$ {\scriptsize(38.5)} & --- & --- & 53.6 {\scriptsize(81.3)} & \textbf{59.7} {\scriptsize(90.7)} & 65.9 \\
\bottomrule
\end{tabular}
\end{table}

\paragraph{Layer agreement for the recompute set.}
\begin{figure}[h]
\centering
\includegraphics[width=0.8\linewidth]{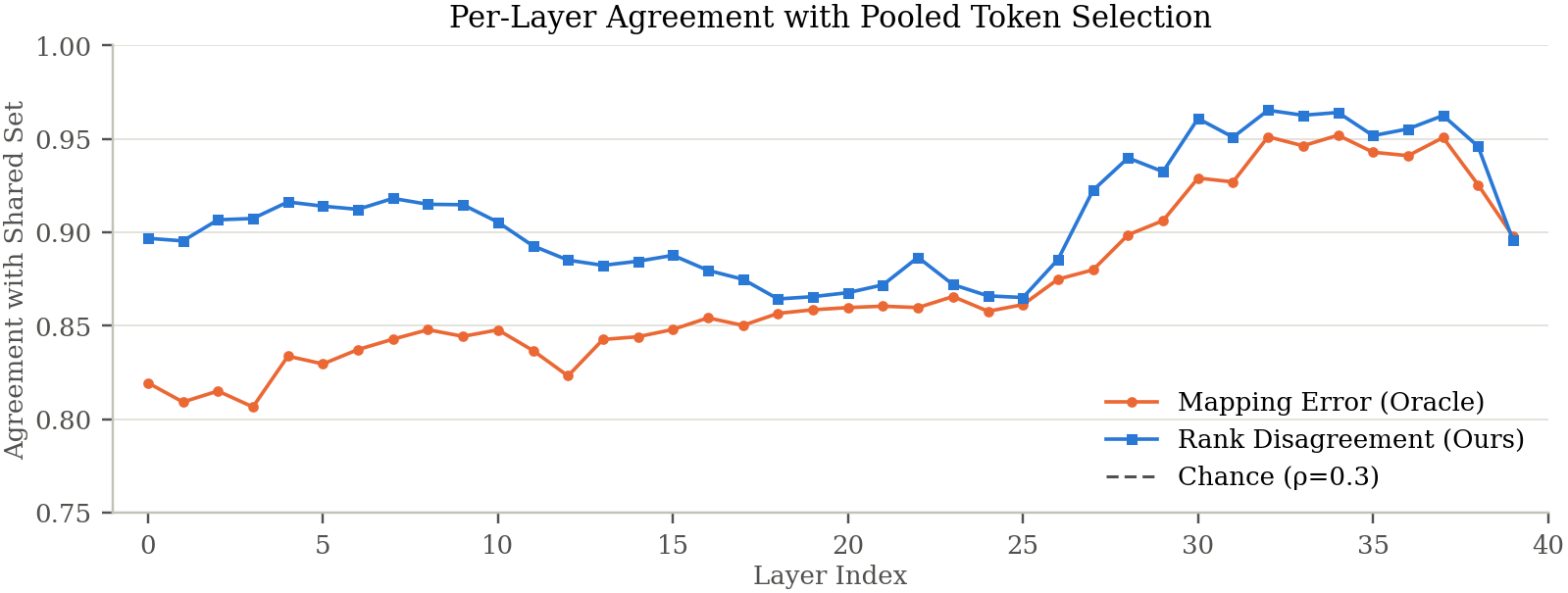}
\caption{Layerwise agreement scores with the shared recompute set determined by $\ell_2$ mapping error and rank disagreement.}
\label{fig:layer-agreement}
\end{figure}
For each target
layer of Qwen3-14B, the overlap between that layer's own top-$\rho$ positions and the
single set the method applies (the top-$\rho$ of the score summed over layers), averaged
over 300 GSM8K prompts, Qwen3-0.6B$\to$14B, $\rho{=}0.3$ is demonstrated in Figure \ref{fig:layer-agreement}. 
Under rank disagreement every layer overlaps
the shared set by 0.86-0.97, and by more than the $\ell_2$ mapping-error oracle does at every
layer but the last. The gap is widest in layers 0-25 (4.8 points on average), where the
oracle's per-layer choices vary most, and narrows to 1.7 points in layers 30-38, where
both scores agree most (0.93-0.97).

\paragraph{Visualizing selected tokens with rank disagreement}
\begin{figure}[t]
    \centering
    \includegraphics[width=\linewidth]{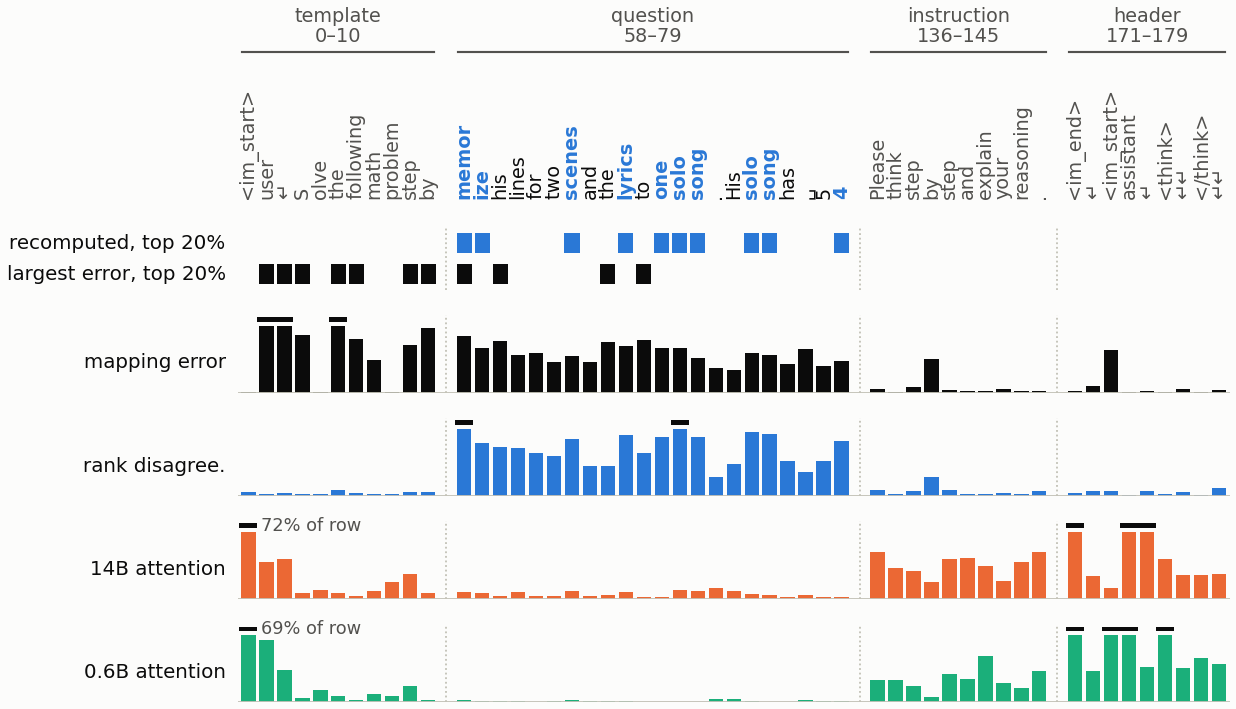}
    \caption{\textbf{What the map gets wrong in one GSM8K prompt, and which tokens each
    signal would recompute.} Columns are prompt
    tokens from four windows, one per section (position ranges above). Tokens in
    \textcolor[HTML]{2A78D6}{\textbf{blue}} are those rank disagreement recomputes at
    $\rho{=}0.2$. The two lanes mark the top 20\% of positions by rank disagreement and by
    mapping error. The four rows below are relative mapping error
    $\lVert\mathrm{mapped}-\mathrm{true}\rVert^2/\lVert\mathrm{true}\rVert^2$ (keys and
    values together, averaged over layers), the rank-disagreement score, and the attention
    each token receives from the 14B and from the 0.6B (last 32 query positions, last quarter
    of layers). Each row is scaled to its own 95th percentile over the whole prompt; a black
    tick marks a clipped value, and the \texttt{<im\_start>} sink's share of its row is
    printed.}
    \label{fig:token-strip-with-ours}
\end{figure}

The Figure \ref{fig:token-strip-with-ours} shows the top 20\% tokens selected with rank disagreement vs. other selector metrics. Rank disagreement selects most of the tokens from the question or content, whereas a large budget for other metrics is spent in template tokens. Recomputing template tokens shows negligible accuracy gains, indicating that the mapper effectively maps the template tokens. Rank disagreement successfully identifies the crucial content tokens helping close the accuracy gap with target prefill.

\section{Limitations and Future Work}
While RaReCache effectively mitigates the accuracy degradation inherent in cross-model transfer, the underlying closed-form linear mapping \citep{heo2026closedform} remains sensitive to the distribution of the calibration data. In our current pipeline, transferring from source to target requires fitting the mapping matrices on a calibration set that closely matches the downstream evaluation distribution (e.g., using the training split of the specific benchmark). Developing generalized, task-agnostic linear maps or curating a highly diverse, universally representative calibration corpus will enable fully zero-shot selective recomputation without prior knowledge of the deployment workload.

RaReCache successfully bridges massive capacity gaps (e.g., $23\times$ from 0.6B to 14B parameters), but it currently assumes that the source and target models share the same tokenizer and vocabulary space, as is standard within model families like Qwen and Llama. Extending this framework to support KV cache transfer and selective recomputation across completely heterogeneous model families (with differing tokenizers) remains an open challenge for future research.


\end{document}